\documentclass[10pt,twocolumn,letterpaper]{article}

\usepackage{cvpr}              % To produce the CAMERA-READY version
\definecolor{cvprblue}{rgb}{0.21,0.49,0.74}
\usepackage[pagebackref,breaklinks,colorlinks,citecolor=cvprblue]{hyperref}

\usepackage{multirow}
\usepackage{xcolor,colortbl}

\usepackage{algorithm}

\usepackage{algpseudocode}

\def\paperID{} % *** Enter the Paper ID here
\def\confName{CVPR}
\def\confYear{2026}

\begin{document}

% ---------------------------------------------------------------

\title{OscNet: Machine Learning on CMOS Oscillator Networks} 

\author{Wenxiao Cai, Thomas H. Lee\thanks{Corresponding Author: tomlee@ee.stanford.edu .} \\ Stanford University
}

\maketitle

\begin{abstract}
Machine learning and AI have achieved remarkable advancements but at the cost of significant computational resources and energy consumption. This has created an urgent need for a novel, energy-efficient computational fabric to replace the current computing pipeline.
Recently, a promising approach has emerged by mimicking spiking neurons in the brain and leveraging oscillators on CMOS for direct computation. In this context, we propose a new and energy efficient machine learning framework implemented on CMOS Oscillator Networks (OscNet).
We model the developmental processes of the prenatal brain's visual system using OscNet, updating weights based on the biologically inspired Hebbian rule. This same pipeline is then directly applied to standard machine learning tasks. 
OscNet is a specially designed hardware and is inherently energy-efficient. 
Its reliance on forward propagation alone for training further enhances its energy efficiency while maintaining biological plausibility.
Simulation validates our designs of OscNet architectures.
Experimental results demonstrate that Hebbian learning pipeline on OscNet achieves performance comparable to or even surpassing traditional machine learning algorithms, highlighting its potential as a energy efficient and effective computational paradigm.
\end{abstract}

\section{Introduction}
\begin{figure*}[h!]
	\begin{center}
    \includegraphics[width=1\linewidth]{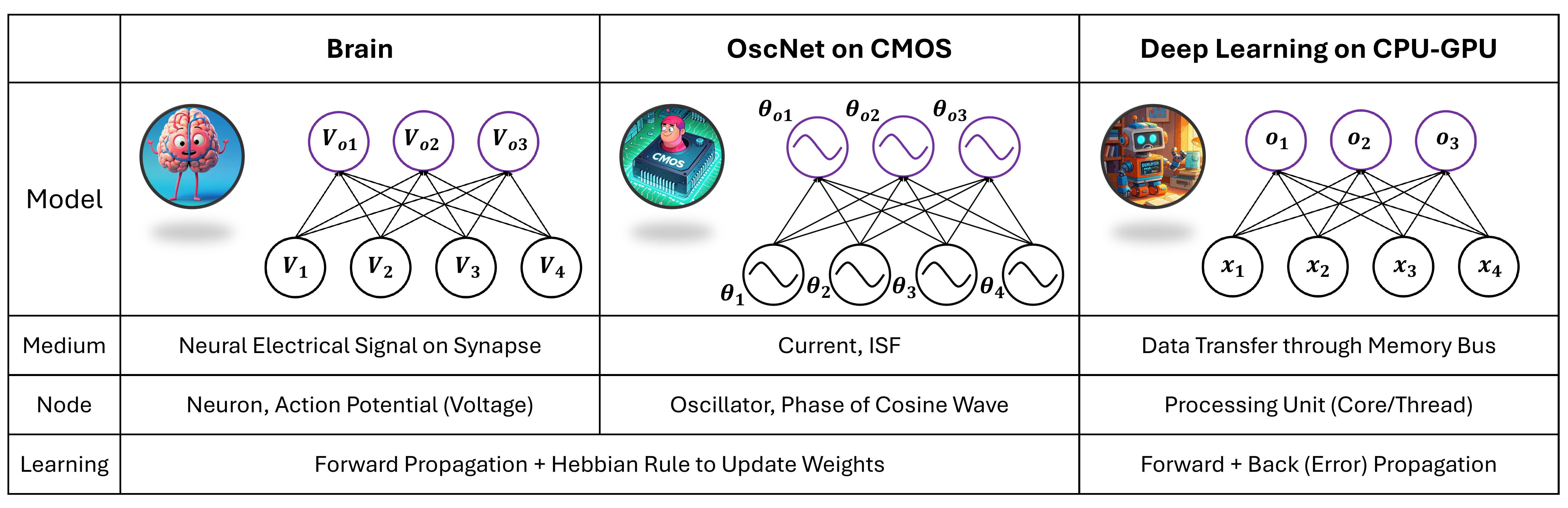}
	\end{center}
\caption{The Oscillator Network (OscNet) is a CMOS circuit inspired by the brain, where current serves as the medium for information transmission between oscillators. The network employs the Hebbian rule for learning, enabling adaptive and efficient connectivity among oscillators. Like the brain, OscNet calculates model responses and update weights with forward propagation only. Compared to recent deep learning models where back propagation is also needed, OscNet saves time and energy.}
\label{fig:teaser}
\end{figure*}

Machine learning, as the cornerstone of Artificial Intelligence (AI), has the capability to tackle a diverse array of significant challenges. 
Its applications~\cite{spatialbot,vdd,objgsp,uncert} span across critical domains such as healthcare~\cite{ml_health_1,ml_health_2,ml_health_3}, finance~\cite{ml_finance_1,ml_finance_2,ml_finance_3}, and industry~\cite{ml_industry_1,ml_industry_2}.
However, with the development of AI, the time and energy costs associated with training models are becoming increasingly high and unsustainable~\cite{train_cost_1,train_cost_2,train_cost_3,train_cost_4}.
To save energy, there is an urgent need for new computational fabrics to replace the existing von Neumann architecture and CPU-GPU-based computation and training pipeline.
Brain-inspired systems have become increasingly promising in recent years~\cite{brain_inspired_1,brain_inspired_2}, particularly Complementary Metal Oxide Semiconductor (CMOS) oscillator systems, which are being explored for optimization architectures~\cite{inject_lock_neuron}. 
These architectures have already been proven to solve some NP-hard problems, such as graph coloring, and can be applied in quantum computing gate emulation~\cite{graph_coloring,quantum_computing}.

In response to this, we propose a new and energy efficient computing and learning pipeline based 
CMOS Oscillator Networks (OscNet). 
OscNet computes with interactions between oscillators and learns with Hebbian rule.
The advantages of OscNet include:
\begin{itemize}
    \item OscNet is a brain-inspired computational architecture and is biologically meaningful.
    In this paper, OscNet is used to simulate the development of the human visual system before birth, specifically the process of establishing connections between retinal cells and lateral geniculate nucleus (LGN);
    \item In machine learning, we propose to update OscNet weights with forward propagation and Hebbian rule, eliminating the computational overhead of back propagation. It has similar or even higher performance with current deep learning pipelines and is energy efficient;
    \item From a hardware perspective, the integrated CMOS OscNet offers the benefits of compact area and compatibility with standard foundries, serving as a promising alternative to the von Neumann architecture.
\end{itemize}

In this paper, we propose the Multi-Input Multi-Output (MIMO) OscNet for image convolution, simulating the development of the human visual system before birth, where retinal cells and LGN neurons establish connections using Hebbian learning rules. 
We find that the Hebbian learning on MIMO OscNet is also suitable for unsupervised machine learning in Auto Encoder fashion, requiring only forward propagation and eliminating the time- and energy-consuming back propagation. 
OscNet is also specially designed to implement K-means.
Furthermore, MIMO OscNet can be used for supervised linear regression. 
In all, the proposed inference and learning pipeline on OscNet is a promising alternative to current framework of machine learning on von Neumann architectures.

\section{Proposed Model}

\subsection{OscNet and Potts Hamiltonian}
CMOS Oscillator Network (OscNet)~\cite{graph_coloring,quantum_computing,inject_lock_neuron} operates with high-order injection locking, with local connections to neighboring oscillators. All oscillators are injected with a master pump signal at $N·f_0$, where $f_0$ is oscillators' natural oscillator frequency, and $N$ is a chosen integral.
The phases of oscillators can be $n \frac{2\pi}{N}$, where $n=1,2...N-1$.
This network mimics the activity of polychronous spiking neurons in the brain~\cite{oscillator_spiking1,oscillator_spiking2}, where each oscillator represents a single brain cell. 
In OscNet, all oscillators oscillate at the same frequency, and the information is encoded in the phases of the individual oscillators.  The voltage across the oscillator $i$ at time $t$ is:
\begin{equation}
    V_i(t)=A_i cos(\omega_0t+\theta_i),
\end{equation}
where $\omega_0$ is the free-running frequency of the oscillator, $A_i$ is the amplitude, and $\theta_i$ is the phase.
% The system can be viewed as oscillators interacting with their neighbors and the main pump, with current serving as the medium of interaction. 
Each oscillator outputs a current based on its own state, while changes in its phase are determined by the influence of neighboring oscillators and the main pump, which can be explained by Impulse Sensitivity Function (ISF) theory~\cite{isf_1,isf_2}.
Phase changes of oscillator $i$ that is connected to neighboring oscillators can be represented by Kuramoto’s equation~\cite{kuramoto_eq,graph_coloring}:
\begin{equation}
    \frac{d\theta_i}{dt} = \sum_{(i,j) \in \mathcal{N}} K_{ij} sin(\theta_i - \theta_j) + K_p sin(N\theta_i),
\end{equation}
where $k_{i,j}$ represents the coupling strength between oscillator $i$ and $j$,  $K_p$ models the main pump current and ISF, and $\mathcal{N}$ is the set of neighboring oscillators of $i$.
The global Lyapunov function exists and will be minimized over time by the oscillator network~\cite{graph_coloring,lyapunov_1,lyapunov_2}: 
\begin{equation}
    E(t) = \frac{N}{2} \sum_{(i,j) \in  \mathcal{N}} K_{ij} cos(\theta_i - \theta_j) + \sum_{i} K_p cos(N\theta_i).
    \label{eq:Lyapunov}
\end{equation}

Eq.~\ref{eq:Lyapunov} takes a similar form of Potts Hamiltonian~\cite{potts_1}:
\begin{equation}
    H = - \sum_{(i,j) \in \mathcal{N}} J_{ij} cos(\theta_i - \theta_j) - \sum_{i} h_i.
    \label{eq:potts}
\end{equation}

Potts Hamiltonian was initially developed to model overall energy and phase transitions in ferromagnetic materials.
The CMOS oscillator network can find the minimal value of Potts Hamiltonian and thus minimize energy of the system. In this paper, we model problems with Potts Hamiltonian, design structures of OscNet and it finds optimal phases of the system.

\subsection{MIMO OscNet}
We propose Multi-Input Multi-Output (MIMO) OscNet, which accepts data inputs and represents the data through the phases of oscillators. In the output layer, OscNet solves the system's Potts Hamiltonian, achieving the same effect as the network's forward propagation in current deep learning frameworks.
The MIMO OscNet is shown in the middle of Fig.~\ref{fig:teaser}. Phases of input oscillators, denoted as $\theta_i$, are fixed, while the output layer oscillators, $\theta_o$, are free-running. 
The main pump of OscNet is $Nf_0$, and we assume that $N$ is infinite in this model. Thus the phases of oscillators can take any value in $[0, 2\pi]$, and $cos(\theta - \theta_i)$ can be anywhere between $-1$ and $1$.
Each output oscillator $\theta_{oj}$ is influenced by the input data $\theta_i$ and the network connection weights, or more specifically, the coupling strength $w_{i,j}$ between the input and output oscillators. Additionally, there is no direct coupling between different output oscillators, such as $\theta_{oj_1}$ and $\theta_{oj_2}$, ensuring they do not interfere with each other.
% We assume that there are infinite number of possible states in this system, so $\theta_i$ and $\theta_o$ can take any value between $0$ and $2\pi$.
Since $J_{ij}$ can be either positive or negative, for simplicity, we ignore the negative sign in Eq.~\ref{eq:potts}.
$h_i$ is a time-independent term and can also be ignored in the differentiation.
Consider an output oscillator $\theta_o$, its value is given by Potts Hamilton:
\begin{equation}
    \theta_{oj} = argmin_{\theta} \: H =  argmin_{\theta} \: \sum_{i=1}^{N} J_{ij} cos(\theta - \theta_i).
    \label{eq:potts_1}
\end{equation}

We igonre $h_i$ in Eq.~\ref{eq:potts_1} as it is independent of $\theta$.
The Potts Hamilton can be solved by:
\begin{align}
        \frac{\partial H}{\partial \theta} &= 0 \\ 
        \frac{\partial H}{\partial \theta} &= -\sum_{i=1}^{N} J_{ij} sin(\theta - \theta_{i}) \\  \nonumber
        &= -\sum_{i=1}^{N} J_{ij}(sin\theta_{i} cos\theta -  cos\theta_{i} sin\theta) \\  \nonumber
        &= -[ cos\theta \sum_{i=1}^{N} J_{ij} sin\theta_{i} - sin\theta \sum_{i=1}^{N} J_{ij} cos\theta_{i} ] \\  \nonumber
        &= \sqrt{(\sum_{i=1}^{N} J_{ij} sin\theta_{i})^2 + (\sum_{i=1}^{N} J_{ij} cos\theta_{i})^2} sin(\theta - \theta_0),  \nonumber
\end{align}

So $\theta_{oj}$ is:
\begin{equation}
    \theta_{oj} = \theta_0 =  arctan \frac{\sum_{i=1}^{N} J_{ij} sin\theta_{i}}{\sum_{i=1}^{N} J_{ij} cos\theta_{i}}.
     \label{eq:potts}
\end{equation}

MIMO OscNet has $N$ input oscillators and $M$ output oscillators. 
It is fully connected so there are $MN$ connections in total.
Since the oscillators in the output layer do not interact with each other, the MIMO OscNet can be viewed as $M$ Multi-input Single-output OscNets sharing the same input but with different coupling strengths, simultaneously solving the Potts Hamiltonian.
In the following chapters, we will demonstrate how MIMO OscNet can model biological development and solve machine learning problems.

\section{Biological Views of OscNet}
\subsection{Image Convolution with OscNet}
In image convolution, $I(x,y)$ and $I_i$ are pixel values at position $(x,y)$ and $i$.  $I'(x,y)$ is pixel value at position $(x,y)$, after convolution.

\begin{equation}
    I'(x,y) = \frac{\sum_{i=1}^{N} w_i I_i }{\sum_{i=1}^{N} w_i }.
    \label{eq:convolution}
\end{equation}

Normalization is applied to Eq.~\ref{eq:convolution}.
Gaussian blurring is a special case of convolution, where $w_i$ is sampled from a 2-dimensional Gaussian distribution.
Considering Eq.~\ref{eq:potts} and Eq.~\ref{eq:convolution} , we map:
\begin{align}
    w_i I_i &= J_i sin\theta_i, \\
    w_i &= J_i cos\theta_i.
\end{align}

So:
\begin{align}
\label{eq:MIMO_convolution_theta}
    \theta_i &= arctan I_i, \\
    \label{eq:MIMO_convolution_J}
    J_i &= \frac{w_i}{cos\theta_i} = w_i \sqrt{I_{i}^{2} + 1}.
\end{align}

OscNet gives us $\theta =  arctan \frac{\sum_{i=1}^{N} J_i sin\theta_i}{\sum_{i=1}^{N} J_i cos\theta_i}$, and $tan\theta$ is $I'(x,y)$ after convolution (or blurring). 
By assigning appropriate coupling weights and phases to the input oscillators, the OscNet can automatically perform image convolution and blurring. The operation of OscNet for image convolution can be succinctly summarized as: given input pixels $I_i = tan\theta_i$, OscNet runs forward propagation once and we can read out from output oscillators that convoluted pixel is $I'=tan\theta$. 
By using multiple output oscillators, we can perform image convolution in parallel.

\subsection{Human Visual System Development}
\label{sec:human_visual_sys}
\begin{figure*}[h]
	\begin{center}
    \includegraphics[width=0.9\linewidth]{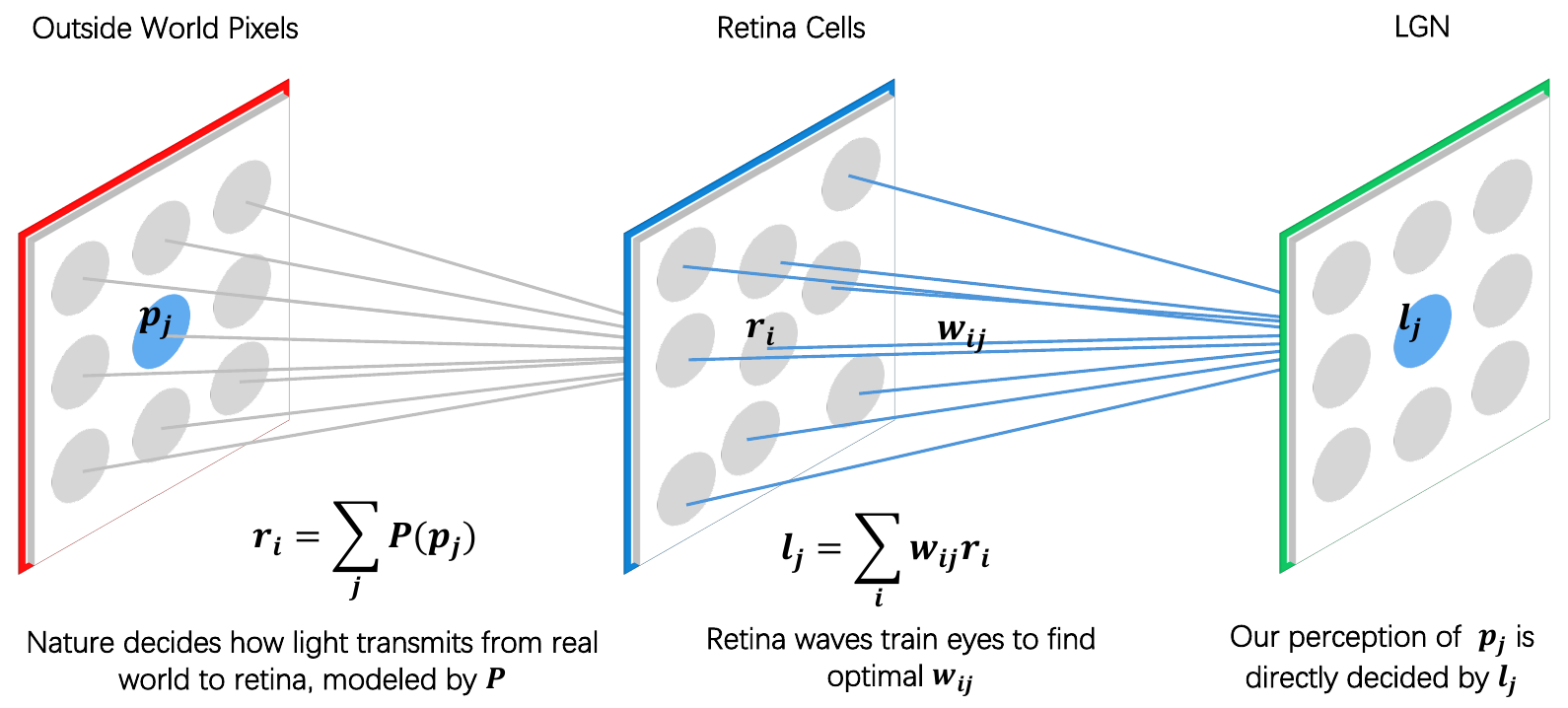}
	\end{center}
\caption{Modeling of human visual perception. Light from the real world travels to the retina, where it is processed by retina cells. These cells are densely connected to the LGN. The retina generates waves that help the brain optimize the weights $w_{ij}$ so that $l_j = p_j$, even without explicitly knowing the positions of the retina cells or having seen any real-world images beforehand.}
\label{fig:retinal_remapping}
\end{figure*}

In this section, we review the mechanisms of visual perception and the development of the visual system before birth, specifically the process of connecting the retina to the lateral geniculate nucleus (LGN)~\cite{visual_develop_1,visual_develop_2,visual_develop_3,visual_develop_4}.
When light enters the retinal cells, it is converted into electrical signals and then transmitted to the brain for processing. 
% This is the basic workflow of the human visual system.
In digital images, pixels are evenly distributed along the x and y axes, but retinal cells are not uniformly distributed on the retina; instead, they are arranged in an irregular pattern. As a result, a straight line in the real world does not project as a straight line onto the retina. \textbf{How do humans perceive a straight line as being straight?}

The modeling of human perception of external light~\cite{retina_grow_nips} is illustrated in Fig.~\ref{fig:retinal_remapping}. 
We model the real world as a set of pixels $p_j$ uniformly distributed along the x and y axes on a plane.
According to the laws of light propagation, light from the outside world falls onto retinal cells, and the values of these retinal cells correspond to the intensity of the light received. 
The value of the retina cell $r_i$ is the summation of real-world pixels, under $P(p_j)$, where $P$ models law of light transmission, effects of lens, and positional relationships between $p_j$ and $r_i$.
Since the brain does not explicitly know the positions of retina cells, we do not explicitly know $P$.
Retinal cells are densely connected to the LGN, where each LGN value represents a linear combination of values from several retinal cells.
We assume that when the brain wants to know the value of pixels in a specific spatial location, it knows which LGN value to read. In other words, there exists a one-to-one mapping between LGN activity and the brain's spatial perception. 
The value of LGN $l_j$ is decided by values of $r_i$ and connections $w_{ij}$ between retina cells and LGN, thus: $l_j =\sum_{i} w_{ij} r_i$.
Therefore, the developmental process of the retinal-LGN connection can be formulated as an optimization problem:
\begin{align}
&\text{Find } w_{ij}, \quad \text{so that } l_j = p_j, \\ \nonumber
&\text{s.t.} \quad l_j = \sum_{i} w_{ij} r_i, \\ \nonumber
&\quad \quad r_i = \sum_{j} P(p_j). \nonumber
\end{align}

Before birth, the visual system undergoes preliminary development. The brain spontaneously generates retinal waves~\cite{retinal_wave_1,retinal_wave_2,retinal_wave_3,retinal_wave_4,retinal_wave_5,retinal_wave_6}, which act as a training signal to establish the correct connections between retinal cells and LGN. As a result, even before experiencing real-world images, the brain already knows that a straight line is straight.
This developmental process can be effectively modeled using OscNet.

\subsection{OscNet Modeling of Visual System}
In the human visual system, transmission of information between retinal cells and the LGN can be seen as image convolution, where pixel values undergo a weighted average. 
Therefore, we can use forward propagation and Hebbian weight updating on OscNet to simulate how humans learn the connection weights between retinal cells and the LGN through retinal waves before birth. 
In OscNet, we use $tan\theta$ to represent numerical values, where $\theta$ is the phase of an oscillator. 
As shown in Algorithm~\ref{alg:hebbian_learning}, the Hebbian learning process involves updating retinal waves, performing forward propagation, and adjusting weights.

\begin{algorithm}
\caption{Hebbian Learning in OscNet with Retinal Waves}
\begin{algorithmic}[1]
\State \textbf{Initialize:}
\State Fully connect retinal cells and LGN oscillators with randomly initialized connection weights.

\For{$i = 1$ to $L$}
    \State (a) Update retinal waves.
    \State (b) Perform OscNet forward propagation to compute LGN responses to retinal waves.
    \State (c) Read out values from LGN oscillators.
    \State (d) Update connection weights according to the Winner-Takes-All Hebbian rule.
\EndFor

\end{algorithmic}
\label{alg:hebbian_learning}
\end{algorithm}

Compared to simulating this process on a von Neumann Architecture (VNA) computer, OscNet requires significantly fewer computational resources. With $n$ retinal cells and $m$ LGN neurons, VNA needs $nm$ computations to forward propagate, whereas OscNet achieves this in a single step.

Compared to deep learning, OscNet more closely aligns with biological development because: 
\begin{itemize}
    \item Forward propagation only: OscNet relies solely on forward propagation, mirroring brain development, whereas deep learning requires both forward and backward propagation;
    \item Local weight updates: In OscNet, weight updates depend only on local information. In biology, synaptic updates are influenced by the activities of the presynaptic and postsynaptic neurons. In contrast, deep learning's weight updates depend on labels and the activities of neurons in higher layers of the network;
    \item Unsupervised learning in early visual system development: In the biological visual system, the retina-LGN pathway develops before birth, long before exposure to real-world images. The brain forms structures capable of perceiving a straight line as a straight line without ever having “seen” one. This process resembles unsupervised learning. In comparison, deep learning typically relies on supervised training.
\end{itemize}

\section{Classification with OscNet}
\subsection{Auto Encoder with Hebbian Learning}
In this section, we use MIMO OscNet for unsupervised Hebbian learning, like an AutoEncoder~\cite{autoencoder_1,autoencoder_2,autoencoder_3}.
AutoEncoders utilize forward and backward propagation to update network weights, compressing input data into a specified dimension for efficient data encoding and representation. In contrast, OscNet requires only forward propagation to update weights and learn data features. 
It achieves results comparable to an AutoEncoder while being more energy-efficient, time-saving, and closely aligned with the brain's natural learning process.
The trained OscNet is then utilized for unsupervised classification and regression tasks. Additionally, we introduce a supervised linear transformation layer to refine the learned representations, following an unsupervised-pretraining and supervised-finetuning fashion.

Similar to Retina-LGN development in Algorithm~\ref{alg:hebbian_learning}, we update OscNet weights with Hebbian rule. MIMO OscNet has $N$ input and $M$ output oscillators with weight matrix $W_{ij}$, where $i\in[1,N]$ and $j\in[1,M]$.
Given training data $(X)$, where $X$ is a feature vector of $N$ components.
We assign input oscillator $i$ to phase $\theta_i = arctan X_i$.
$X$ is forward propagated, so output oscillator $j$ will be 
\begin{equation}
    \theta_j = arctan \frac{\sum_{i=1}^{N}W_{ij}X_{i}}{\sum_{i=1}^{N}W_{ij}}.
\end{equation}

We choose $j^*$ so that:
\begin{equation}
    j^* = argmax_j tan\theta_j.
\end{equation}

According to Hebbian rule, when an axon of cell A is near enough to excite a cell B and repeatedly or persistently takes part in firing it, some growth process or metabolic change takes place in one or both cells such that A’s efficiency, as one of the cells firing B, is increased~\cite{hebbian_book}.
Winner-Takes-All (WTA) strategy is adopted.
We increase connections between the maximum $tan j$  and input oscillators, while other connections fade. The overall weight updating strategy is:

\begin{equation}
    \text{Update}(W_{ij}) = 
\begin{cases} 
W_{ij} + \tan\theta_{j^*}(X-\tan\theta_{j^*}W_{ij}), &  j = j^*, \\
W_{ij} - \lambda \tan\theta_{j^*}(X-\tan\theta_{j^*}W_{ij}), & j \neq j^*,
\end{cases}
\end{equation}

where $\lambda$ is a coefficient that controls the rate at which the weights decay. In this way, OscNet learns features in the dataset in a unsupervised manner that aligns with brain development principles.
The learned OscNet is ready for unsupervised classification and regression.
Additional, OscNet can be used to train an additional linear regression layer in a supervised manner. This linear layer performs a weighted combination of the learned features to improve prediction accuracy. Fine-tuning involves only the linear regression layer, resulting in minimal parameters and low resource consumption.
This approach of unsupervised pretraining followed by supervised finetuning~\cite{pretrain_finetune_1,pretrain_finetune_2} is widely adopted in modern deep learning frameworks.

\subsection{OscNet K-means}
Given input $x_{i}$ and network weight $w_{ij}$, the output of OscNet $y_j$ can be written as:
\begin{equation}
\label{eq:hebbian_kmeans_oscnet}
    y_j = \sum_{i}x_{i}w_{ij} = \mathbf{w}_j \cdot \mathbf{x}.
\end{equation}

For OscNet Hebbian learning in clustering, a data point is clustered to class $j^*$ if $y_{j^*} = max_j y_j$. According to Eq.~\ref{eq:hebbian_kmeans_oscnet}, we can define the overall energy term (or error) as the relative number of cosine similarity~\cite{hebbian_kmeans}.
Thus the center of a class $C_j$ is ideally:
\begin{equation}
     \mathbf{w}_j = \sum_{\mathbf{x} \in C_j} \frac{\sum \mathbf{x} }{\|\mathbf{x}\|}.
\end{equation}

In OscNet, we use weight $\mathbf{w}_j$ to represent center of cluster $C_j$.
Given input data $\mathbf{x}$ and current network weight $\mathbf{w}$, forward propagation of OscNet finds the cluster $\mathbf{x}$ belongs to by checking the maximum output oscillator. 
Thus we can use OscNet for K-means.
OscNet only needs to run once to determine which cluster the data belongs to. Compared to calculating distances multiple times on a CPU, it saves both time and energy.

\section{Regression with OscNet}
In this section, we describe supervised learning using OscNet. Specifically, we apply OscNet to linear regression, where an analytical solution exists. Since nonlinear regression problems can be transformed into linear ones using polynomial transformations~\cite{poly_regression_1,poly_regression_2}, SVM~\cite{svm_regression_1,svm_regression_2,svm_regression_3}, and kernels~\cite{kernel_regression_1,kernel_regression_2,kernel_regression_3}, the proposed OscNet can be applied to a broader range of regression problems.

\subsection{Single Variable Linear Regression}
We start with a single variable object function.
The training data is denoted as $(x_i, y_i)$, $i = 1, 2, ... N$ where $N$ is the number of training data. We aim to fit $f(x) = kx$ on the training data. The objective can be written as minimizing Mean Squared Error (MSE) loss:
\begin{equation}
    l(k) = \sum_{i=1}^{N} (kx_i - y_i)^2 .
    \label{eq:single_variable_regression}
\end{equation}

To find $k = argmin_{k} l(k)$, we compute the derivative of $l(k)$:
\begin{align}
    \frac{\partial l(k)}{\partial k} &= \frac{2}{N}  \sum_{i=1}^{N} (x_i^2k - x_i y_i) = 0.
\end{align}

Thus:
\begin{equation}
    k = \frac{ \sum_{i=1}^{N} x_i y_i}{ \sum_{i=1}^{N} x_i^2}.
    \label{eq:min_loss_single_parameter}
\end{equation}

Comparing Eq.~\ref{eq:min_loss_single_parameter} and Eq.~\ref{eq:potts}, we cast the regression problem to Potts Hamilton by:
\begin{align}
    x_i y_i &= J_i sin\theta_i, \\
    x_i^2 &= J_i cos\theta_i.
\end{align}

This gives:
\begin{align}
\label{eq:single_var_regression_weight}
    \theta_i &= arctan \frac{y_i}{x_i}, \\
    \label{eq:single_var_regression_weight_2}
    J_i &= \frac{x_i^2}{cos\theta_i} = x_i \sqrt{x_i^2 + y_i^2}. 
\end{align}

We can set the coupling strength following Eq.~\ref{eq:single_var_regression_weight} and Eq.~\ref{eq:single_var_regression_weight_2}, so OscNet can find the solution $k$ to linear regression, according to Eq.~\ref{eq:min_loss_single_parameter}

\subsection{Multivariable Regression with Coordinate Descent}
We focus on multivariable linear regression with coordinate descent. 
Coordinate Descent minimizes loss by iteratively minimizing a multivariable function along one coordinate direction at a time while keeping other variables fixed, which is commonly used in machine learning and statistics for solving regression problems.
Please note that a lot of regression problem can be rewritten into linear regression. For example,  $\theta_1 x^2 + \theta_2 x$ can be rewritten into $\theta_1 x_1 + \theta_2 x_2$. In kernel regression, we use kernel $\phi(x)$ and solve $f(\phi(\mathbf{x})) = \sum_{i=1}^{N}(\theta_i \phi_i(\mathbf{x})) + \theta_0$. It is also important to note that linear regression is a single-layer neural network in deep learning.

On training data $(\mathbf{x_i}, y_i) = (x_{i1}, x_{i2}, ..., x_{iM}, y_i)$ where $i = 1,2...N$, we aim to fit $f(\mathbf{\theta},\mathbf{x}) = \theta_0 + \sum_{k=1}^{M} \theta_k x_k $. 
The MSE loss function is:

\begin{equation}
    l(\mathbf{\theta}) = l(\theta_0, \theta_1, \theta_2, ... \theta_m) = \sum_{i=1}^{N} ( \theta_0 + \sum_{k=1}^{M} \theta_k x_{ik} - y_i)^2.
\end{equation}

In coordinate descent, we choose $\theta_j$, fix every other $\theta_k$ ($k \neq j$), and find $\theta_j$ to minimize $l(\mathbf{\theta})$. So loss can be rewritten into:
\begin{equation}
    l(\mathbf{\theta}) = l(\theta_j) = \sum_{i=1}^{N} (\theta_j x_{ij} - h_i)^2,
    \label{eq:multi_variable_regression}
\end{equation}

where $h_i = y_i - \theta_0 - \sum_{k=1, i \neq j}^{M} \theta_k x_k $ is a constant, given that we fix every $\theta_k (k \neq j)$ and know $(\mathbf{x_i}, y_i)$ from training data.
Eq.~\ref{eq:multi_variable_regression} is exactly the same with single variable regression Eq.~\ref{eq:single_variable_regression}. In coordinate descent, we iteratively choose $j$ and update$\theta_j$ until the loss function converges to a locally optimal value.

\section{Simulations}
To verify the effectiveness of OscNet at the circuit level, we simulate MIMO OscNet using MATLAB~\cite{matlab}, based on Kuramoto's equation~\cite{kuramoto_eq}.
Given the network structure, coupling strength, and the phase of the input oscillator, we observe the phase of the free-running oscillators, which serves as the output.

\begin{figure}[h]
	\begin{center}
    \includegraphics[width=1\linewidth]{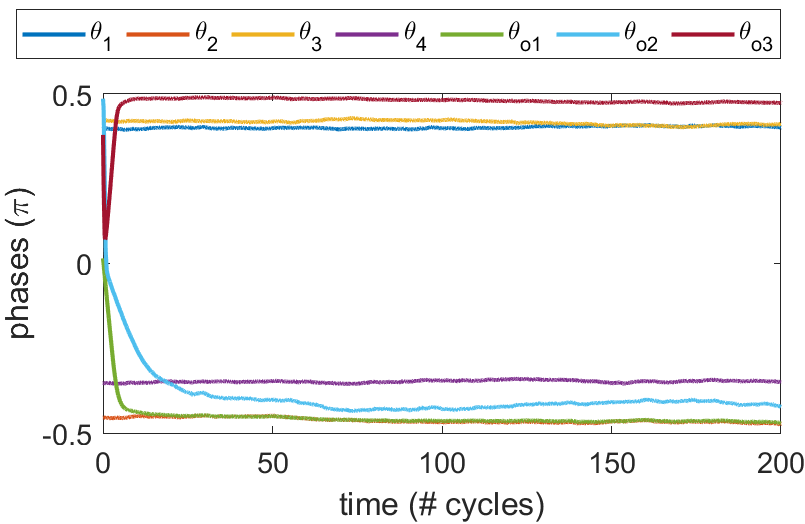}
	\end{center}
\caption{MIMO OscNet simulation for 4 inputs and 3 outputs.}
	\label{fig:mimo-simulation}
\end{figure}

We simulate MIMO OscNet with $N=4$ input oscillators and $M=3$ outputs.
Each row represents an input oscillator $\theta_{i}$, and each column is an output oscillator $\theta_{j}$. 
Input numerical values are $X = [3,-7,4,-2]$, and encoded to $\theta = [0.397\pi, -0.455\pi, 0.422\pi, -0.352\pi ]$ following Eq.~\ref{eq:MIMO_convolution_theta}.
The weights matrix is:
\begin{equation}
W = 
\begin{bmatrix}
-3 & -9 & 0 \\
8 & 2 & -6 \\
-1 & 7 & 10 \\
7 & 5 & -1
\end{bmatrix},
\end{equation}

In OscNet, the weights of oscillator connections are encoded following Eq.~\ref{eq:MIMO_convolution_J}. As shown in Fig.~\ref{fig:mimo-simulation}, OscNet finds phases of output oscillators to be $\theta_o = [-0.458\pi, -0.432\pi, 0.489\pi]$, which corresponds to $O = [-7.545, -4.600, 28.000]$. The simulation results agree with the theory.

\section{Experiments}
In the previous section, we validated the feasibility of MIMO OscNet. In this section, we use an unsupervised pretrained OscNet to perform unsupervised tasks followed by supervised fine-tuning.

\subsection{Unsupervised Classification}
\begin{figure}[h]
	\begin{center}
    \includegraphics[width=1\linewidth]{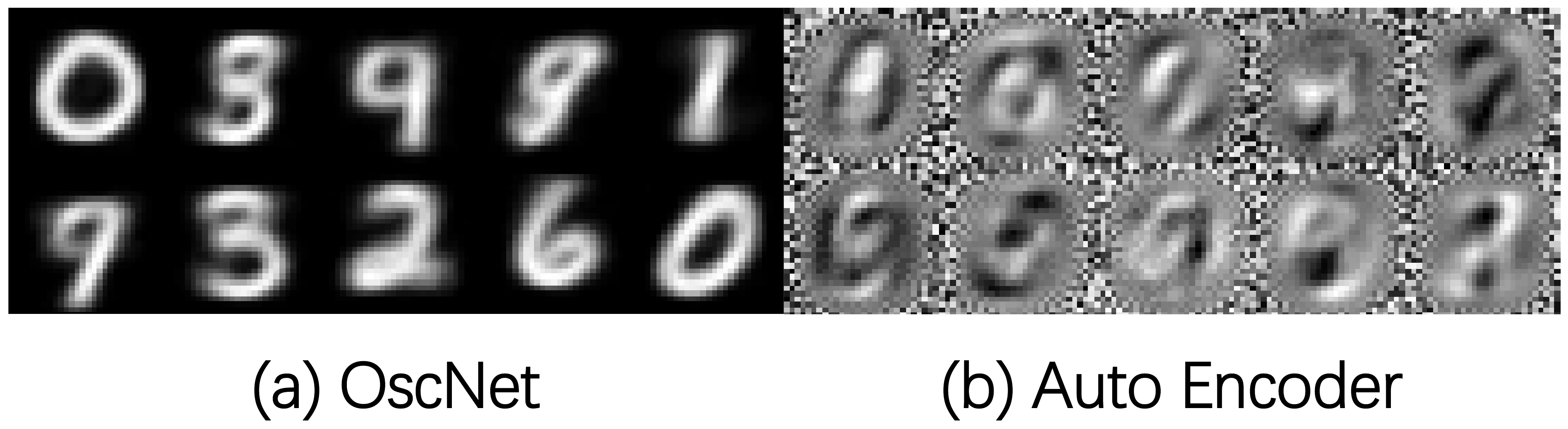}
	\end{center}
\caption{Learned features of OscNet and Auto Encoder on MNIST with 10 hidden neurons.}
	\label{fig:learned_features}
\end{figure}

We use the handwritten digit classification dataset MNIST. The number of input features is $N = 784$, and the output has $M = 10$ classes. We train the network using the Hebbian rule to update the weights. During the testing phase, given an input image, after forward propagation, if output oscillator $j$ exhibits the maximum response, we classify the image as belonging to class $j$. 
As a comparison, we also use an AutoEncoder~\cite{autoencoder_1,autoencoder_2,autoencoder_3}. The AutoEncoder has $N$ input nodes, $M$ hidden nodes in one layer, and $N$ output nodes, trained using error propagation. However, the performance of a single hidden-layer AutoEncoder is suboptimal. Therefore, we also include a two-hidden-layer AutoEncoder for comparison, where the two hidden layers have 128 and 10 nodes, respectively. 
Experimental results are shown in Table.~\ref{table-unsupervised-mnist}. OscNet with Hebbian learning reaches a much higher accuracy than AutoEncoders.
The learned features of OscNet and AutoEncoder are shown in Fig.~\ref{fig:learned_features}.
Please note that this is just a fully connected network, without using priors such as convolutional kernels. The receptive field corresponding to each hidden neuron is autonomously learned during training.

\begin{table}[h]
\caption{Accuracy of OscNet and AutoEncoders on MNIST test set. Models are all trained on MNIST train set in a unsupervised manner.}
\centering

\begin{tabular}{lc}

\toprule
Model & Accuracy (\%) \\
\hline
AutoEncoder - 1 Hidden Layer & 13.77 \\
AutoEncoder - 2 Hidden Layer & 28.72 \\
\textbf{OscNet (Ours)} & \textbf{57.31} \\

\bottomrule
\end{tabular}
\label{table-unsupervised-mnist}
\end{table}

\subsection{Supervised Finetuning}

After unsupervised pretraining of OscNet, we perform supervised training of a linear regression layer on the learned features and test the classification accuracy on MNIST.
Experimental results are shown in Table.~\ref{table-supervised-ft-mnist}.
Compared to the deep learning pipeline where an Auto Encoder-Auto Decoder structure is pretrained and a linear regression layer is fine-tuned on the AutoEncoder in a supervised manner, OscNet excels at more extreme data compression. With fewer hidden neurons, OscNet achieves higher accuracy, demonstrating superior compression and feature extraction. When the hidden layer size increases, OscNet achieves results comparable to those of the Auto Encoder. This suggests that OscNet, as a more energy-efficient pipeline, has the potential to replace existing machine learning frameworks without compromising accuracy or performance.

\begin{table}[h]
\caption{OscNet and Hebbian rule pretraining, followed by finetuning a linear regression layer on MNIST. Auto Encoder pretraining - linear regression finetuing is the baseline, which is trained on error back propagation.}
\centering

\begin{tabular}{lcc}

\toprule
Model & \# Hidden Neurons & Accuracy (\%) \\
\hline
AutoEncoder & 10 & 73.84 \\
\textbf{OscNet (Ours)} & 10 & \textbf{77.47} \\
\hline
AutoEncoder & 16 & 81.22 \\
\textbf{OscNet (Ours)} & 16 & \textbf{85.61} \\
\hline
AutoEncoder & 64 & 89.96 \\
\textbf{OscNet (Ours)} & 64 & \textbf{90.58} \\
\hline
AutoEncoder & 128 & \textbf{91.35} \\
\textbf{OscNet (Ours)} & 128 & 91.33 \\

\bottomrule
\end{tabular}
\label{table-supervised-ft-mnist}
\end{table}

\subsection{OscNet K-means}
The experimental results of OscNet K-means is shown in Table.~\ref{table-kmeans}. OscNet with cosine similarity reaches almost same performance than K-means with Euclidean distance error implemented on CPU.

\begin{table}[h]
\caption{K-means compared with OscNet K-means on MNIST.}
\centering

\begin{tabular}{lc}

\toprule
Model  & Accuracy (\%) \\
\hline
K-means & 59.46 \\
\textbf{OscNet K-means} & \textbf{60.84} \\

\bottomrule
\end{tabular}
\label{table-kmeans}
\end{table}

\section{Related Work}
\subsection{Polychronous Oscillatory Networks}
The Ising model~\cite{ising_1,ising_2,ising_3,ising_4} was discovered in the context of ferromagnetism in statistical mechanics. The Ising Machine leverages the natural tendency of such systems to minimize their energy, enabling it to efficiently find optimal solutions~\cite{ising_solve_1,ising_solve_2,ising_solve_3,ising_solve_4}. It has been widely applied to various combinatorial optimization problems. In an Ising Machine, each node can only exist in one of two possible states.
In this paper, we build upon the principles of CMOS oscillator networks and the Ising Machine to design a system capable of converging to the minimum of the Potts Hamiltonian, where each node can have multiple states. Polychronous Oscillator Networks on CMOS are inspired by spiking neurons, mimicking their interactions, and have been successfully applied to NP-hard problems such as Graph Coloring, as well as in designing quantum computers.
Inspired by neural computing~\cite{neural_comp_1,neural_comp_2,neural_comp_3,neural_comp_4} and their implementation on hardware~\cite{circuit_learn_1,circuit_learn_2,circuit_learn_3}, we design a MIMO OscNet structure, where some oscillators serve as inputs and others as outputs. The hardware can autonomously perform inference, effectively enabling forward propagation. The primary goal of this paper is to design an appropriate learning strategy to complement this hardware, enabling its application in machine learning tasks.

\subsection{Hebbian Learning}
Biological systems learn through signal forward propagation and Hebbian learning~\cite{hebbian_bio_1,hebbian_bio_2}. In essence, Hebbian theory states that an increase in synaptic efficacy arises from the repeated and persistent stimulation of a postsynaptic cell by a presynaptic cell~\cite{hebbian_synaptic_1,hebbian_synaptic_2}.
Before birth, humans undergo initial learning of the visual system spontaneously. Retinal waves~\cite{retinal_wave_1,retinal_wave_2,retinal_wave_3,retinal_wave_4,retinal_wave_5,retinal_wave_6,retinal_wave_7} are generated on the retina as input signals, and the connection weights between retinal cells and the LGN are updated based on Hebbian theory~\cite{retina_grow_nips}. 
Similar learning mechanisms, rooted in Hebbian learning~\cite{hebbian_math_1}, are also applied in principal component analysis (PCA)~\cite{hebbian_pca}, sparse coding~\cite{hebbian_sparse_coding}, reinforcement learning~\cite{hebbian_control_1, hebbian_control_2} and unsupervised learning in neural networks~\cite{hebbian_unsupervised_1,hebbian_unsupervised_2,hebbian_unsupervised_3,hebbian_unsupervised_4, hebbian_unsupervised_5,hebbian_unsupervised_6}. Using the Winner-Takes-All (WTA) strategy, network weights are updated efficiently.
In this paper, we model the early visual system development using Hebbian theory and OscNet to address the question: How do humans perceive a straight line as being straight in Sec.~\ref{sec:human_visual_sys}. We design a pipeline where forward propagation and Hebbian weight updating serve as the foundation for OscNet's unsupervised learning. This pipeline demonstrates its applicability not only in unsupervised learning tasks but also in general supervised machine learning tasks.

\section{Conclusion}
In this paper, we propose OscNet, an energy-efficient machine learning framework based on CMOS oscillators.
OscNet leverages interactions of oscillators to achieve minimization of Potts Hamiltonian. 
Compared to traditional deep learning architectures, OscNet more closely mirrors the structure of the human brain while being significantly more energy-efficient.
We introduce a phase-based representation for numerical values in OscNet, applying it to machine learning tasks where solving the Potts Hamiltonian corresponds to forward propagation. 
Inspired by Hebbian learning in the human brain, we use OscNet to model the development of the prenatal human visual system and extend this model to unsupervised network pre-training.
A trained OscNet can perform unsupervised classification in Auto Encoder or K-means fashion or be fine-tuned in a supervised manner with an additional layer. 
These promising features suggest that OscNet has the potential to become a computational fabric for next-generation AI systems.

\clearpage  

\bibliographystyle{splncs04}
\bibliography{main}
\end{document}